\documentclass{styles/svproc}
\usepackage[hyphens]{url}
\usepackage{tcolorbox}

\usepackage{graphicx}
\usepackage{mathrsfs} 

\begin{document}
\mainmatter              
\title{Cat \& Mouse -- Can Fake Text Generation Outpace Detector Systems?}
\titlerunning{Cat \& Mouse}  
%
\author{Andrea McGlinchey \inst{1} \and Peter J Barclay\inst{2} 
}
\authorrunning{McGlinchey \& Barclay} 

\institute{Lumerate, Canada \\
email: andrea.mcglinchey@lumerate.com \\
\and
School of Engineering, Computing, and the Built Environment \\
Edinburgh Napier University, Scotland \\
email: p.barclay@napier.ac.uk}

\maketitle              

\begin{abstract}
Large language models (LLMs) can produce convincing `fake text' in domains such as academic
writing, product reviews, and political news. Many approaches have been investigated
for the detection of artificially generated text. While this may seem to presage an
endless `arms race', we note that newer LLMs use ever more parameters, training data, 
and energy, while relatively simple classifiers  demonstrate a good level of 
detection accuracy with modest resources.
To approach the question of whether the models ability to beat the detectors may therefore
reach a plateau, we examine the ability of statistical classifiers to identify 
`fake text' in the style of classical detective fiction. Over a 0.5 version increase,
we found that Gemini showed an increased ability to generate deceptive text, while
GPT did not. This suggests that reliable detection of fake text may remain feasible even
for ever-larger models, though new model architectures may improve their deceptiveness.

\end{abstract}

\section{Background}

With the rise of sophisticated text generation models, there has been an increase in the volume of fake news, social media posts 
and other generated text used to spread disinformation \cite{koplin}. 
Owing to the ease of misuse of Large Language Models (LLMs) by bad actors, it is crucial that detectors are developed to identify AI-generated content and inauthentic materials.

Misuse of LLMs is well studied, with work on detection methods completed in the fields of fake news, student submissions, user contributions and medical texts \cite{crothers}; however, in comparison, very few studies have focused on creative writing. The growing importance of this area was recently highlighted by authors protesting against Meta in opposition of the use of a database containing millions of copyrighted books to train its AI models \cite{creamer}.

So far, only two studies have addressed this domain. 
The ``Ghostbuster'' study \cite{verma} used three domains for data preparation, one being creative writing,
which proved slightly harder to detect. The datasets were created using GPT-3.5-turbo, and it achieved an F1 score of 98.4\% for the creative writing data set. The
``AI Detective'' study \cite{mcglinchey} focused solely on creative writing, and also using GPT-3.5-turbo,  the highest model available at that time. The F1 scores for this system ranged from 94.2\% to 100\%, outperforming Ghostbuster on shorter extracts. The authors hypothesised that as LLMs continued to be 
created with greater numbers of parameters (weights) and trained with ever more data, the effectiveness of detectors might decrease.  

\section{Problem Statement}

With the rapid evolution of LLM tools, we might expect that it is becoming ever-harder to identify artificially generated text. This 
could lead to a ``cat \& mouse'' game where generators and 
detectors improve incrementally, with neither gaining a clear
advantage.

However, 
earlier work such as \cite{mcglinchey} and \cite{wahde} has shown
that classical classifiers can reach a good level of accuracy, and we note
that such algorithms are subject to relatively little resource constraint.
LLMs, on the other hand, are employing vast numbers of parameters, with huge training sets, 
and exorbitant energy demands \cite{jegham}. We speculate therefore that their ability to fool well
trained classical detectors may plateau over time.
This paper describes our early attempts to address this Research Question. 

\begin{tcolorbox}
Do later versions of the same LLM show an increased 
ability to generate deceptive text, resulting
in decreased accuracy of the detector?
\end{tcolorbox}

\section{Methodology}

Creative fiction can be generated by an LLM either from scratch, 
following a prompt, or by re-writing human-authored text; whilst both these methods are addressed in the studies cited,  
here we focus on the rewrite method for greater comparability between different models. This approach would also be relevant
for detecting paraphrased text which could be used for
other forms of plagiarism.

Datasets were created using six out-of-copyright novels by Agatha Christie. The novels were chunked into approximately 100 word excerpts, with full stops  used as separators. The excerpts were then randomised and sent through OpenAI and Gemini's API to be rewritten. It was requested that the re-written text should be approximately the same length as the input text. 
The prompt had been optimised on OpenAI, and was kept the same for Gemini for consistency. For Open AI, three models were tested -- the oldest available model GPT 3.5-turbo (GPT 3.5), the next model up which has been optimised for speed and cost GPT 4o-mini, and the latest available model GPT 4.1. Two models were tested in Gemini, 1.5-flash model (Gemini 1.5) and 2.0-flash model (Gemini 2.0). 

As highlighted in earlier work \cite{mcglinchey}, it is difficult to get the LLMs to generate
texts of the required lengths; while the average length of a human text was 574, GPT 3.5 achieved an average of 502, GPT 4o-mini achieved 603 and finally GPT 4.1  hit an average of 581. The Gemini models gave much shorter text excerpts with an average of 404 for version 1.5 and 412 for version 2.0. 
Therefore, after generation the data sets were pruned to balance them with approximately the same average length, median length and standard deviation, to allow a fair comparison of human-written
and AI-generated text. Table \ref{tab:lengthcomp} shows the balanced data set metrics.

Finally, each data set was randomised and split to give 80\% for training and validation, retaining 20\% for an unseen test set. For each trial, 80\% of the data was used to create a balanced, randomised dataset of mixed human-written and AI-generated texts, and this was
used to train each of the
four machine-learning classifiers: Support Vector Machine; Random Forest; Naive Bayes and an MLP Classifier.
The remaining 20\% was then used to create a balanced, mixed test set and used to calibrate the accuracy of each classifier.

\begin{table}[h]
\caption{Comparing statistics for text chunks}
\label{tab:lengthcomp}
\centering
\begin{tabular}{|l|r|r|r|r|}
\hline
\textbf{Dataset} & \textbf{No. Rows} & \textbf{Mean Len.} & \textbf{Median Len.} & \textbf{Std. Dev.} \\ \hline
Original Human Text & 2713 & 574.31 & 578 & 65.80 \\ \hline
Balanced Human Text, GPT  & 1735 & 579.70 & 580 & 43.33 \\ \hline
AI Text, GPT 3.5 & 2713 & 502.21 & 515 & 118.67 \\ \hline
Balanced AI Text, GPT 3.5 & 1735 & 571.41 & 563 & 67.57 \\ \hline
AI Text, GPT 4o-mini & 2713 & 602.86 & 605 & 78.58 \\ \hline
Balanced AI Text, GPT 4o-mini & 1735 & 578.21 & 583 & 40.53 \\ \hline
AI Text, GPT 4.1 & 2713 & 580.93 & 583 & 77.79 \\ \hline
Balanced AI Text, GPT 4.1 & 1735 & 580.70 & 582 & 70.95 \\ \hline
Balanced Human Text, Gemini  & 1000 & 509.19 & 523.5 & 47.92 \\ \hline
AI Text, Gemini-1.5 & 2713 & 404.07 & 400 & 73.64 \\ \hline
Balanced AI Text, Gemini-1.5 & 1000 & 479.97 & 470 & 44.51 \\ \hline
AI Text, Gemini-2.0 & 2713 & 411.98 & 410 & 77.19 \\ \hline
Balanced AI Text, Gemini-2.0 & 1000 & 490.98 & 480 & 47.41 \\ \hline
\end{tabular}

\label{tab:dataset_statistics}
\end{table}

\section{Results}

The generated fragments were first eyeballed to confirm their similarity to the human-written texts.
Table \ref{tab:wording} shows an example of one human text and how each LLM rewrote it. 
We noted that the fragments generated by
Gemini are generally shorter and the layout of the rewrites are similar to each other. The GPT rewrites  seem to mostly keep the same sentence structure as the human excerpts, and use synonyms to swap out words. 
On inspection, the increasing versions of the GPT models do not noticeably produce different types of rewrites; however, it is possible to see the improvements  between Gemini 1.5 and 2.0. 

Comparing the results shown in Table \ref{tab:modelcomp}, Gemini 2.0 consistently proved
most deceptive -- where the false results were mostly  AI-generated text misclassified as human -- with a
noticeable increase of over 10\% between version 1.5 and 2.0. 
(For brevity we report only Accuracy, which was almost the same as the F1 scores over our balanced datasets).

Unexpectedly, and in contrast to Gemini, GPT 4.1 did not produce more deceptive texts than version 3.5.
Less surprisingly, text from GPT-4o-mini was easier to detect, with the MLP Classifier reaching almost 98\% accuracy. This model was designed for speed and efficiency, and may be expected to be
less capable than GPT versions 3.5 and 4.1.

\begin{table*}[htbp]
\tiny
\centering
\caption{Comparison of same human excerpt re-written by different models}
\label{tab:wording}
\begin{tabular}{| p{1cm} | p{9cm} |}
\hline
\textbf{Text Preparation}   &  \textbf{Text Excerpt} \\
\hline
\textbf{Original Human Text}   & His face, clean-shaven and exquisitely mobile, was stamped with an expression of power and force far beyond the ordinary. Magnetism seemed to radiate from him. Tuppence was undecided for the moment whether to put him down as an actor or a lawyer, but her doubts were soon solved as he gave her his name: Sir James Peel Edgerton. She looked at him with renewed interest. This, then, was the famous K.C. whose name was familiar all over England. She had heard it said that he might one day be Prime Minister. \\
\hline
\textbf{Model GPT 3.5}   & Tuppence observed his striking face, impeccably smooth and remarkably expressive, exuding an aura of dominance and intensity that surpassed the norm. Uncertain if he was a thespian or legal professional, her inquiry was swiftly answered when he introduced himself as Sir James Peel Edgerton. Instantly captivated, Tuppence recognized him as the renowned K.C. whose reputation spanned the entirety of England, with rumors suggesting he could potentially ascend to the position of Prime Minister in the future. \\
\hline
\textbf{Model GPT 4o-mini}   & His visage, impeccably groomed and remarkably expressive, bore an aura of authority and strength that surpassed the mundane. Charisma appeared to emanate from his presence. Tuppence found herself momentarily uncertain whether to classify him as an actor or a barrister, but her uncertainty quickly dissipated when he introduced himself: Sir James Peel Edgerton. Her interest piqued anew. This was indeed the renowned K.C., a name recognized throughout England. Rumors had circulated that he could potentially ascend to the position of Prime Minister. \\
\hline
\textbf{Model GPT 4.1} & His face was smooth and expressive, marked by a commanding presence that set him apart from most men. There was a distinct energy about him, almost magnetic in its intensity. At first, Tuppence hesitated, uncertain if he was a member of the stage or the bar, but her uncertainty vanished when he introduced himself: Sir James Peel Edgerton. Instantly, she regarded him with heightened curiosity. This, after all, was the renowned King's Counsel known throughout England. Rumor even suggested he could someday become Prime Minister. \\
\hline
\textbf{Model Gemini 1.5} & Undecided, Tuppence eyed Sir James Peel Edgerton. His exquisitely mobile, clean-shaven face radiated power and force, far beyond the ordinary.  An actor? A lawyer?  His name solved her doubts.  Famous K.C., his name echoed throughout England.  Prime Minister, some said.  She looked at him with renewed interest; this man possessed magnetism. \\
\hline
\textbf{Model Gemini 2.0}   & The noted K.C. Sir James Peel Edgerton, whose famous name echoed throughout England, faced her. Tuppence hesitated. Was he lawyer or actor? Power and force radiated from him. His clean shaven face, exquisitely mobile, held magnetism beyond the ordinary. One day, she thought, he might even be Prime Minister. This explained the renewed interest she now had.  \\
\hline    
\end{tabular}
\end{table*}

\begin{table*}[htbp]
\caption{LLM text generation -- comparison of accuracy using four models
}
\label{tab:modelcomp}
\centering
\begin{tabular}{| l | r | r | r | r |}
\hline
\textbf{LLM Used} & \textbf{Rand. Forest} & \textbf{SVM } & \textbf{MLP Classifier} & \textbf{Naïve Bayes} \\
\hline
Gemini 1.5 & 95.25\% & 96.25\% & 97.00\% & 96.25\%\\
\hline
Gemini 2.0 & 82.25\% & 85.50\% & 83.00\% & 79.75\%\\
\hline
GPT 3.5 & 89.48\% & 92.94\% & 94.09\% & 94.81\%\\
\hline
GPT 4o-mini & 92.36\% & 96.83\% & 97.84\% & 97.26\%\\
\hline
GPT 4.1 & 90.78\% & 93.95\% & 95.24\% & 94.38\%\\
\hline
\end{tabular}
\end{table*}

\begin{table*}[htbp]
\caption{LLM comparison using Deception Rate, $\mathscr{D} = 1 - $Recall}
\label{tab:modelcompdrate}
\centering
\begin{tabular}{| l | r | r | r | r |}
\hline
\textbf{LLM Used} & \textbf{Rand. Forest} & \textbf{SVM } & \textbf{MLP Classifier} & \textbf{Naïve Bayes} \\
\hline
Gemini 1.5 & 6.00\% &   3.00\% &    3.50\% &    14.00\%\\
\hline
Gemini 2.0  & 24.50\% & 14.00\% &   18.00\% &   34.00\%\\
\hline
GPT 3.5 & 17.29\% & 9.22\% &    9.22\% &    8.93\%\\
\hline
GPT 4o mini & 11.82\% & 3.46\% &    2.31\% &    4.61\%\\
\hline
GPT 4.1 Balanced & 14.12\% &    6.63\% &    6.05\% &    9.22\%\\
\hline
\end{tabular}
\end{table*}

To compare the models more intuitively, we introduce a new metric Deception Rate, defined as $\mathscr{D} = 1 - $Recall; this is the proportion of
AI-generated texts which are misclassified as human-written. The Deception Rate for each 
model against each classifier is shown in Table \ref{tab:modelcompdrate}.
The SVM classifier performed well (suffered lower deception rates)
for the all models overall, with the MLP classifier also performing well against 
the GPT models.

\section{Discussion}

We must be careful comparing the GPT and Gemini models to each other directly, as they have different architectures; while both GPT and Gemini use a Transformer architecture,
Gemini uses a Mixture of Experts approach, and has optimisations designed to
support a longer context window. A good overview can be found in \cite{rahman}. 
We used the flash versions for all experiments with Gemini, and note that these proved significantly 
more deceptive than the distilled GPT-4o-mini model, which may expected to be broadly
similar in `size'.

Overall, the Gemini models proved more effective at generating human-like text,
at least within the domain of classic detective fiction, even though the prompt
we used had been previously optimised for GPT models.
Our results suggest that the GPT models may now be hitting a plateau in their ability to
generate deceptive text.

Incidentally, we note that the GPT models do show improvement in 
their ability to generate texts of a requested
length, while the Gemini models are less able to do so, showing minimal improvement across versions.

While model parameters are not publicly disclosed, GPT-4 is believed to have 1 --  2 trillion parameters (weights) compared to GPT-3.5's 157 billion. This means that with a tenfold increase in the model's `size', it is not significantly better at generating text.

The Gemini flash models are distilled models, based on the Gemini-pro models which likely have a comparable number of parameters to the GPT models. On the assumption that there is
a significant increase in the number of parameters between Gemini 1.5 and 2.0, we see that this does result in an increased ability to fool the detectors, with accuracy dropping by over 10\% between versions.
We therefore speculate that the Gemini architecture is better suited
to this task, and better able to benefit from an increase
in size of the model.

\section{Conclusion and Further Work}

Our results
show that for Gemini models, the version increase from 1.5 to 2.0 shows a significant increase
in the model's ability to generate convincing text in the style of classic detective novels; for the GPT models,
however, the version increase shows no significant improvement. This suggests that the GPT models may be reaching a plateau in terms of their ability to generate deceptive text, and that model architecture may become more important than the size of the model. 

This study is a first attempt to investigate the likely future trajectory of LLMs' ability
to generate human-like text.
Further research is hampered by non-disclosure of model parameters, and the unavailability of earlier versions of 
current models. An alternative line of research would be to evaluate a self-hosted model such as NanoGPT or variants of LLaMA, where the number of parameters could be increased incrementally and compared with the model's ability to beat the detectors.
Whether future versions of Gemini can continue to increase in
their deceptive ability remains to be seen; however, 
our results to date call into question the presumption that LLMs will become ever-better
at generating deceptive text.

%
%

\end{document}